\documentclass{article}

\newif\ifarxiv
\arxivtrue

\usepackage{natbib}
\usepackage[utf8]{inputenc} 
\usepackage[T1]{fontenc}    
\usepackage{url}            
\usepackage{booktabs}       
\usepackage{amsfonts}       
\usepackage{nicefrac}       
\usepackage{microtype}      
\usepackage{xcolor}         
\usepackage{fullpage}
\usepackage{kpfonts}
\usepackage{xcolor}

\usepackage[textsize=tiny]{todonotes}
\usepackage{pifont}
\usepackage{multirow} 

\usepackage{colortbl}  
\usepackage{color}
\usepackage{xcolor}
\usepackage{array}   
\usepackage{caption}
\usepackage{float}

\definecolor{DarkGreen}{rgb}{0.1,0.5,0.1}
\definecolor{DarkRed}{rgb}{0.5,0.1,0.1}
\definecolor{DarkBlue}{rgb}{0.1,0.1,0.5}
\definecolor{Gray}{rgb}{0.2,0.2,0.2}
\usepackage[pdftex]{hyperref}
\hypersetup{
    unicode=false,          
    pdftoolbar=true,        
    pdfmenubar=true,        
    pdffitwindow=false,      
    pdfnewwindow=true,      
    colorlinks=true,       
    linkcolor=DarkBlue,          
    citecolor=DarkGreen,        
    filecolor=DarkRed,      
    urlcolor=DarkBlue,          
    pdftitle={update the title},
    pdfauthor={update the authors},
}

\usepackage{subcaption}
\usepackage{graphicx}

\title{Towards Compatible Fine-tuning for Vision-Language
Model Updates}
\author{
Zhengbo Wang$^{1,2}$, Jian Liang$^{2,3}$ \thanks{Correspondence to: Jian Liang (liangjian92@gmail.com)}, 
Lijun Sheng$^{1,2}$,
Ran He$^{2,3}$,
Zilei Wang$^{1}$,
Tieniu Tan$^{2,3,4}$ \\
$^{1}$ University of Science and Technology of China \\
$^{2}$ NLPR \& MAIS, Institute of Automation, Chinese Academy of Sciences \\
$^{3}$ School of Artificial Intelligence, University of Chinese Academy of Sciences \\
$^{4}$ Nanjing University \\
\texttt{zhengbowang@mail.ustc.edu.cn, liangjian92@gmail.com} \\
}
\date{}

\begin{document}
\maketitle

\begin{abstract}
So far, efficient fine-tuning has become a popular strategy for enhancing the capabilities of foundation models on downstream tasks by learning plug-and-play modules. 
However, existing methods overlook a crucial issue: \textit{if the underlying foundation model is updated, are these plug-and-play modules still effective?}
In this paper, we first conduct a detailed analysis of various fine-tuning methods on the CLIP in terms of their compatibility with model updates.
The study reveals that many high-performing fine-tuning methods fail to be compatible with the upgraded models.
To address this, we propose a novel approach, \textbf{C}lass-c\textbf{on}di\textbf{t}ioned \textbf{Co}ntext \textbf{Op}timization (\textbf{ContCoOp}), which integrates learnable prompts with class embeddings using an attention layer before inputting them into the text encoder.
Consequently, the prompts can dynamically adapt to the changes in embedding space (due to model updates), ensuring continued effectiveness.
Extensive experiments over 15 datasets show that our ContCoOp achieves the highest compatibility over the baseline methods, and exhibits robust out-of-distribution generalization.
\end{abstract}

\section{Introduction}

\label{intro}

In the current era, foundation models~\citep{radford2021learning, kenton2019bert, brown2020language, rombach2022high, caron2021emerging} have emerged as the cornerstone of the field of deep learning.
Through pre-training on exceptionally large datasets, these models demonstrate remarkable zero-shot capabilities and generalization, rendering them extensively employed across various domains.

Efficient fine-tuning has emerged as a prominent area of research in the context of large foundation models~\citep{zhou2022learning, zhou2022conditional, hu2021lora, li2021prefix, houlsby2019parameter}. 
By freezing the parameters of foundation models, these approaches train lightweight plug-and-play modules to quickly and cost-effectively adapt the model to downstream tasks, such as learning residual matrices~\citep{hu2021lora, dettmers2023qlora} or additional learnable prompts~\citep{zhou2022learning, li2021prefix, houlsby2019parameter}.
Consequently, by maintaining a frozen large model alongside corresponding lightweight modules, we can low-costly apply the foundation model to thousands of downstream tasks.

The existing efficient fine-tuning methods neglect a crucial problem. 
To improve model performance or attain a safety alignment~\citep{ouyang2022training, sun2023eva, xu2023wizardlm, touvron2023llama, vicuna2023, rombach2022high},  the foundational models at the core are often updated, such as the transition from GPT-3 to ChatGPT, CLIP to EVA-CLIP, and the series of versions in stable diffusion.
However, prior to model updates, we have trained a variety of plug-and-play modules on the current version of the foundation model. 
This raises a significant question:
\begin{center}
  \emph{\textit{
  Can these efficient fine-tuning modules be compatible \\with the upgraded foundation model?}}
\end{center}
Given the substantial costs associated with retraining numerous plug-in modules, exploring the compatibility of these modules emerges as a meaningful pursuit.

\noindent
\begin{minipage}{0.48\textwidth}
\hspace{1em}
We investigate the compatibility of efficient fine-tuning methods on vision-language models (VLMs) in the context of model upgrades.
Such methods can be categorized into two types: one involves adding learnable prompts at the model's shallow layers, while the other entails incorporating learning modules to refine the text features at the model's deep layers.
The latter methods generally exhibit superior performance when applied to downstream tasks.

\hspace{1em}
We first investigate the compatibility of existing methods.
We train these plug-and-play modules on CLIP and directly integrate them into the upgraded model, EVA-CLIP.
The results reveal an interesting observation: while deep-layer fine-tuning methods demonstrate superior performance, their compatibility to updated models falls short, even performing worse than the zero-shot performance of the upgraded model.
Consequently, we posit that, for the model upgrade of VLMs, the shallow layers exhibit better transferability compared to the deeper layers.
\end{minipage}
\hspace{0.02\textwidth}
\begin{minipage}{0.48\textwidth}
    \centering
    \includegraphics[width=1.0\textwidth]{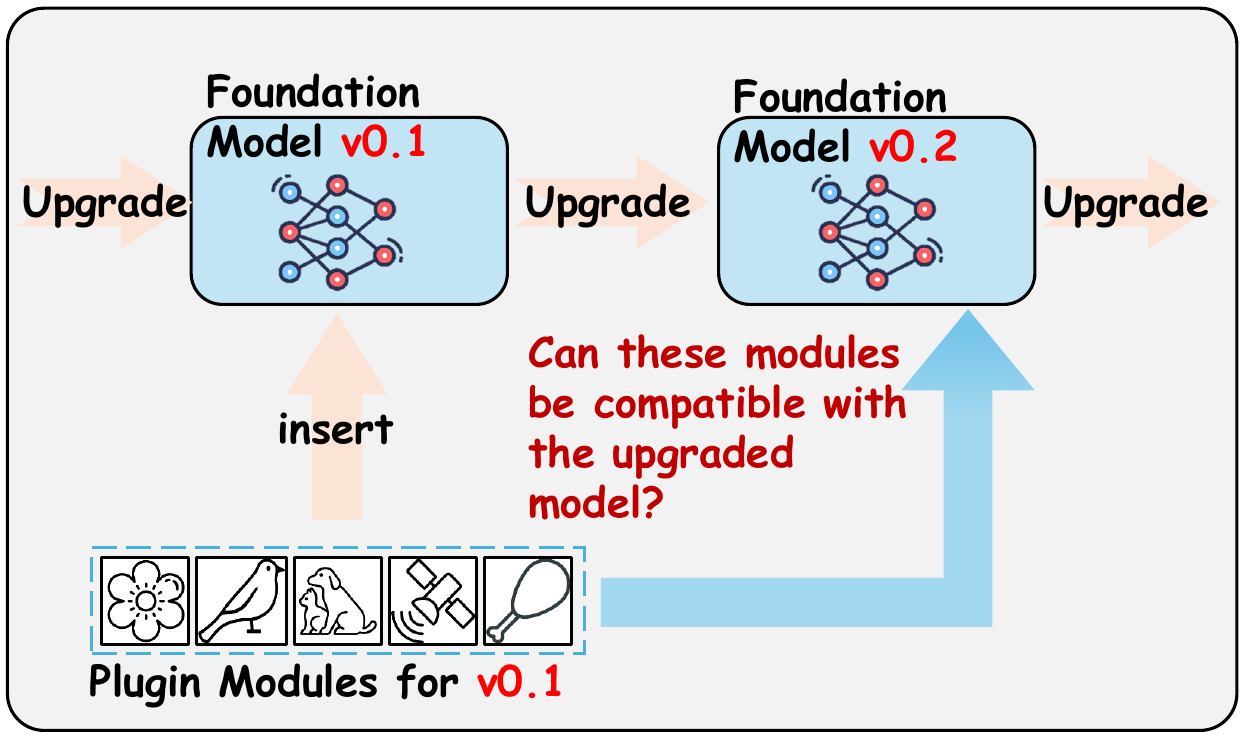}
    \captionof{figure}{
    Efficient fine-tuning methods enable us to easily train plug-and-play modules to enhance the performance of foundation models. 
    However, a significant challenge arises due to the frequent updates in foundational models, such as variants of Llama and CLIP. 
    Re-training these modules for the upgraded model incurs significant costs.
    Therefore, our investigation aims to address the question of \textit{whether these modules can be compatible with the upgraded model}.
    }
    \label{fig:setup}
\end{minipage}

To delve deeper into this phenomenon, we compute the average absolute and relative changes between parameters and output features for each layer before and after model upgrades. 
The analysis reveals that the changes in shallower layers are consistently smaller, confirming the better transferability of these layers. 
To address potential concerns related to methodological influences, we conduct additional experiments. 
Specifically, we train CoOp~\citep{zhou2022learning} at different layers.
The results consistently show a decline in reused performance with increasing layer depth, providing empirical support for our hypothesis.

To address the problem, we propose a novel approach, \textbf{C}lass-c\textbf{on}di\textbf{t}ioned \textbf{Co}ntext \textbf{Op}timization (\textbf{ContCoOp}), designed to enhance compatibility.
Our approach involves learning class-conditioned prompts at the input of the text encoder, akin to shallow-layer methods.
Prior to inputting into the text encoder, we integrate class information into the learnable prompts through an attention network.
Consequently, these prompts possess the capability to dynamically evolve alongside updates to the class embedding within the enhanced model. 
This dynamic adaptability significantly enhances the compatibility of the prompts in upgraded models.

Our contributions are summarized as follows:
\begin{itemize}
    \item To the best of our knowledge, this paper is the first work exploring the module compatibility problem in VLMs.
    Given the high cost associated with re-training the fine-tuning modules after model updates, investigating fine-tuning methods with strong compatibility in upgraded models becomes imperative.
    \item We assess the compatibility of existing efficient fine-tuning methods in VLMs for model upgrades.
    Our study reveals an interesting observation: shallower layers demonstrate superior transferability compared to deeper layers. 
    This discovery can potentially enhance the design of fine-tuning methods in the future.
    \item We propose a novel approach named \textbf{C}lass-c\textbf{on}di\textbf{t}ioned \textbf{Co}ntext \textbf{Op}timization (\textbf{ContCoOp}).
    Leveraging an attention network, we obtain class-conditioned prompts for downstream tasks, allowing for dynamic updates to accommodate the improved model, thereby enhancing compatibility.
    \item We conduct extensive experiments over 15 datasets, which shows that our method exhibits superior compatibility compared to the baseline methods.
\end{itemize}

\begin{figure*}[tbp]
    \centering
    \begin{minipage}{1.0\textwidth}
    \centering
    \setlength{\tabcolsep}{6pt}
    \resizebox{1.0\textwidth}{!}{
        \begin{tabular}{c|c|ccc|ccc}
        \toprule
              & \textbf{ZS-CLIP} & \textbf{LP$\dagger$}    & \textbf{CLIP-A$\dagger$} & \textbf{Tip-A$\dagger$} & \textbf{CoOp}  & \textbf{CoCoOp} & \textbf{KgCoOp} \\
          & (ICML'21) & (ICML'21) & (IJCV'23) & (ECCV'22) & (IJCV'22) & (CVPR'22) & (CVPR'23) \\
        \midrule
        Base  & 65.51  & 78.15  & 79.56  & \textbf{81.63}  & 79.94  & 76.27  & 78.30  \\
        New   & 70.79  & 1.83  & 67.30  & 69.63  & 75.40  & 75.39  & \textbf{75.77}  \\
        H     & 68.04  & 3.57  & 72.92  & 75.15  & \textbf{77.60}  & 75.83  & 77.01  \\
        \bottomrule
        \end{tabular}%
     }
    \end{minipage}
    \begin{minipage}{0.35\textwidth}
        \includegraphics[width=1.0\textwidth]{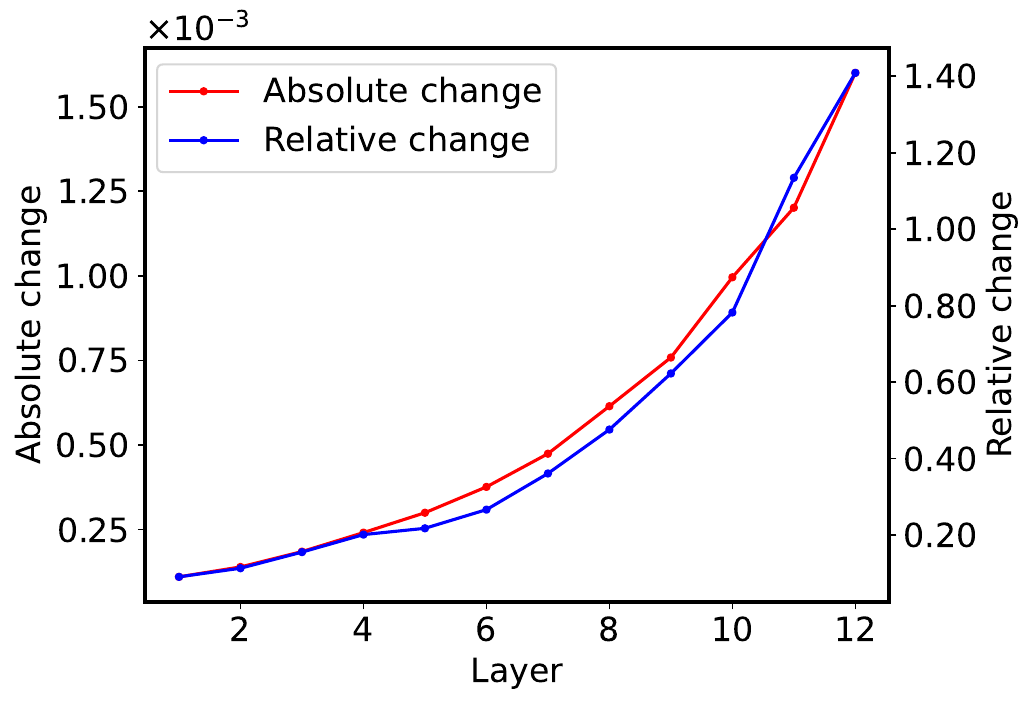}
    \end{minipage}
    \begin{minipage}{0.33\textwidth}
        \includegraphics[width=1.0\textwidth]{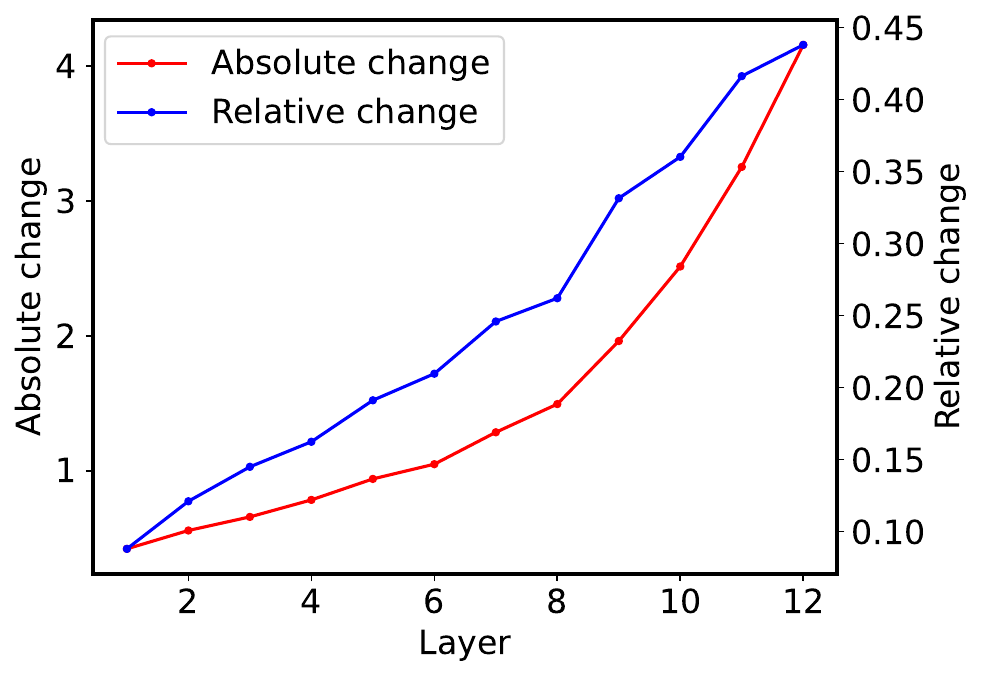}
    \end{minipage}
    \begin{minipage}{0.3\textwidth}
        \includegraphics[width=1.0\textwidth]{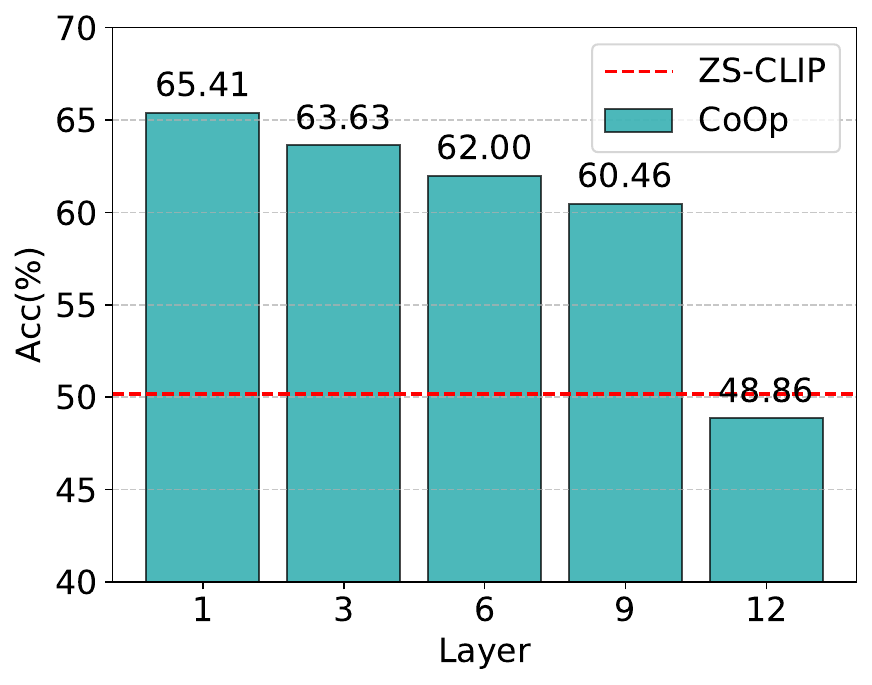}
    \end{minipage}
    \captionof{figure}{
    (a) The average performance over 11 datasets.
    We assess the compatibility of efficient fine-tuning methods of VLMs for model upgrades.
    These methods are trained on ViT-B/16-based CLIP and then integrated into the corresponding EVA-CLIP, the upgraded model.
    The terms Base and New represent the performance tested after inserting these modules into CLIP and EVA-CLIP, with H indicating their harmonic average.
    $\dagger$ denotes that is a deep-layer method.
    (b) The average absolute and relative changes in parameters at each layer of the text encoder before and after model upgrading. 
    (c) The average absolute and relative changes in output features at each layer of the text encoder before and after model upgrading.
    (d) To mitigate methodological impacts, we train CoOp at different layers on the DTD dataset and report its accuracy on the upgraded model. 
    The results indicate that shallower layers exhibit superior transferbility compared to deeper layers.
 }
    \label{fig:preliminary}
\end{figure*}

\section{Related Work}

\label{related}
\noindent
\textbf{Vision-Language Models.}
Vision-Language Models (VLMs) have emerged as a new kind of foundation model, aiming to connect vision and language modalities~\citep{radford2021learning, kim2021vilt, lu2019vilbert, su2019vl, jia2021scaling, sun2023eva, cherti2023reproducible, li2023clipa}.
Trained on extensive image-text datasets, these models showcase remarkable zero-shot recognition and generalization capabilities and they have been widely applied in open-world recognition~\citep{joseph2021towards, gu2021open, liang2023open} and text-to-image generation~\citep{rombach2022high, ramesh2021zero, ramesh2022hierarchical}.
In this paper, our focus is on CLIP~\citep{radford2021learning} and its upgraded version, EVA-CLIP~\citep{sun2023eva}.
Both models are composed of an image encoder and a text encoder, trained with a contrastive loss to align representations from different modalities.
EVA-CLIP leverages the pre-trained text encoder of CLIP and undergoes further contrastive learning on an improved visual model, ultimately showcasing superior performance.

\noindent
\textbf{Efficient Fine-Tuning for VLMs.}
To efficiently and cost-effectively adapt models to downstream tasks, recent research has concentrated on the development of effective fine-tuning methods within VLMs~\citep{zhou2022learning, zhou2022conditional, wang2023improving, gao2023clip, zhang2022tip, yao2023visual, wang2024baseline, shu2022test}.
These methods learn small plug-and-play modules that enhance the model by seamlessly integrating into the frozen model.
Broadly, these methods can be categorized into two types based on the insertion position: shallow-layer and deep-layer methods.
Shallow-layer methods~\citep{zhou2022learning, zhou2022conditional,wang2023improving}, such as CoOp~\citep{zhou2022learning}, focus on learning effective and robust prompts concatenating in the initial layer of the CLIP text encoder.
In contrast, deep-layer methods~\citep{zhang2022tip, gao2023clip, wang2024baseline} learn to refine the output features with an adapter network or a cache model.
While deep-layer methods typically exhibit superior performance on downstream tasks, our observation reveals that, in the context of adapting to updated models, shallow-layer methods demonstrate better compatibility.

\noindent
\textbf{Model Upgrades.}
While foundation models exhibit strong performance and have been applied across various domains, continuous upgrades are essential to enhance their performance and achieve safety alignment~\citep{touvron2023llama, rombach2022high, sun2023eva, ouyang2022training}.
In the field of NLP, model updates are ubiquitous. GPT-3, for instance, has evolved to ChatGPT through instruction fine-tuning and reinforcement learning with human feedback.
The Llama series has undergone numerous upgrades through training on various datasets, resulting in versions such as Alpaca~\citep{alpaca}, Vicuna~\citep{vicuna2023}, LLaVa~\citep{liu2023visual}, and more. 
Similarly, within the realm of generative models, frequent model upgrades are commonplace, exemplified by different versions of Stable Diffusion~\citep{rombach2022high}.
In the domain of VLMs, model upgrades remain prevalent. 
For instance, CLIP~\citep{radford2021learning} has been upgraded to the corresponding EVA-CLIP~\citep{sun2023eva}. 
However, preceding these model upgrades, we have trained several plug-and-play fine-tuning modules. 
Retraining these modules after model updates incurs high costs. 
Due to limitations in computational resources, this paper predominantly investigates the module compatibility challenges in VLMs.
We aim to devise an efficient fine-tuning method that exhibits robust compatibility for model updates in VLMs.

\section{Method}
\begin{figure*}[t]
    \centering
    \includegraphics[width=1.0\textwidth]{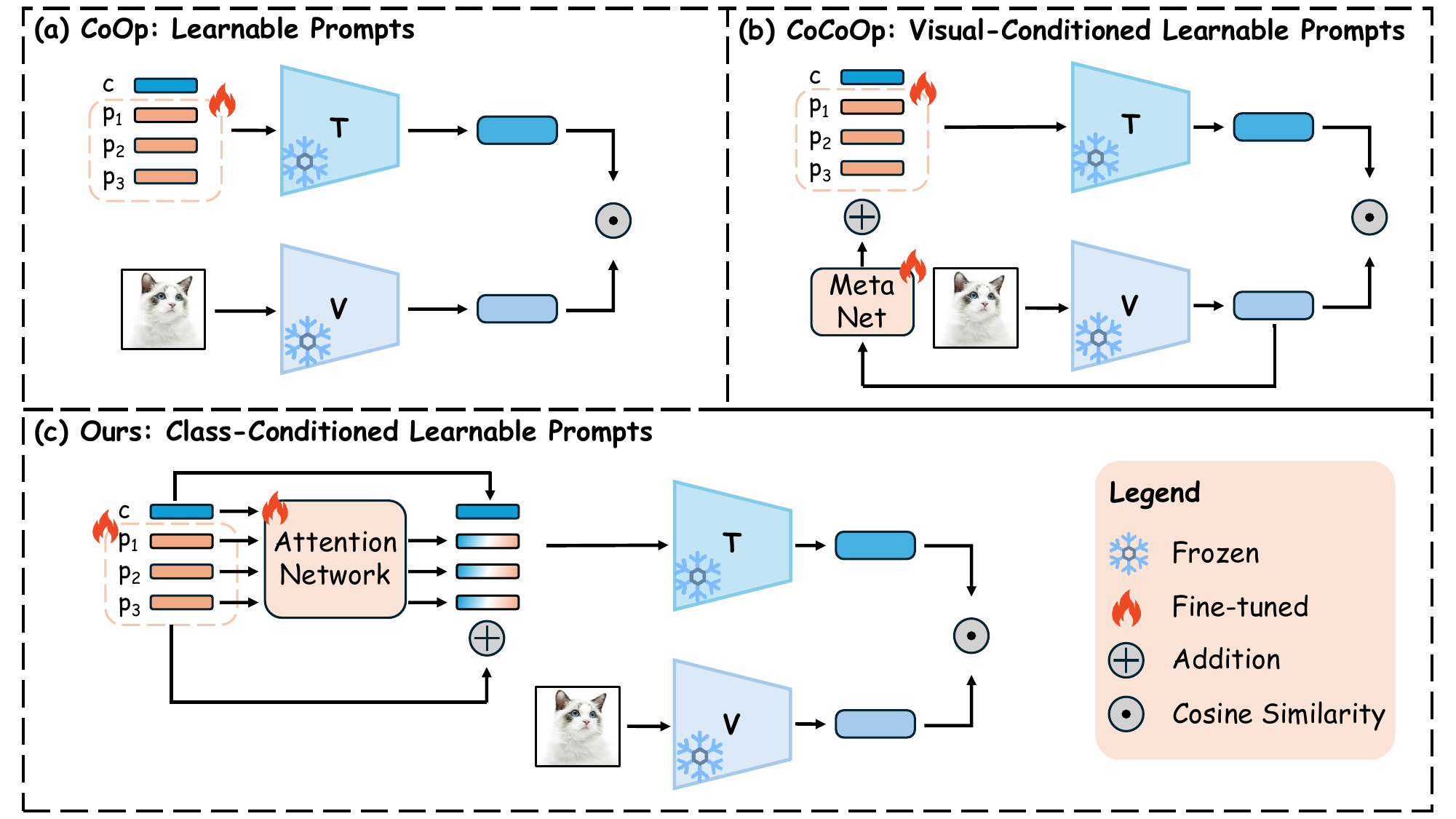}
    \captionof{figure}{
    \textbf{The overview of our method.}
    To enhance the compatibility, we aim for our modules to dynamically adapt to model updates.
    For this, we adopt class-conditioned learnable prompts.
    Leveraging the attention network, our method integrates class information into learnable prompts. 
    Moreover, following model updates, the prompts also undergo automatic updates, synchronized with changes in class embeddings.
    We include CoOp and CoCoOp for comparison.
 }
    \label{fig:architecture}
\end{figure*}

\subsection{Preliminary Study}
In this paper, we concentrate on investigating the compatibility of efficient fine-tuning modules in VLMs for model updates. 
These efficient fine-tuning models are trained on the CLIP model, and we aim to explore their compatibility with the updated model, EVA-CLIP.

\noindent
\textbf{CLIP \& EVA-CLIP.}
CLIP~\citep{radford2021learning} and EVA-CLIP~\citep{sun2023eva} are contrastive-based vision language models, comprising an image encoder $I$ and a text encoder $T$ for encoding inputs from their respective modalities.
As these models have been trained on large-scale image-text pairs, we can easily obtain the zero-shot classifier weight $\{w_i\}_{i=1}^K$ through prompting, where $w_i = T(P_i)$, and $P_i$ is the text prompt such as ``a photo of a [class]" for $i$-th class.
EVA-CLIP, an upgraded version of CLIP, leverages the text encoder from CLIP and further aligns it with an improved image encoder through training on image-text pairs. It exhibits superior performance compared to CLIP.

\noindent
\textbf{Can existing methods be compatible with the upgraded model?}
First, we investigate the compatibility of existing efficient fine-tuning methods for model updates.
We assess the compatibility of six baselines: Linear Probe~\citep{radford2021learning}, CLIP-Adapter~\citep{gao2023clip}, Tip-Adapter~\citep{zhang2022tip}, CoOp~\citep{zhou2022learning}, CoCoOp~\citep{zhou2022conditional}, and KgCoOp~\citep{yao2023visual}.
The first three are deep-layer methods, while the latter three are shallow-layer methods.
We train these plug-and-play modules on ViT-16/B-based CLIP and evaluate their performance when integrated into CLIP and EVA-CLIP.
We denote their performance on CLIP and EVA-CLIP as base and new, respectively, and report the harmonic mean (H) to balance these two metrics.

We report their average performance over 11 datasets at Figure~\ref{fig:preliminary} (a).
We have observed an intriguing phenomenon: the methods inserted into shallow layers exhibit superior compatibility compared to those inserted into deep layers, despite the latter typically demonstrating better performance on downstream tasks.
Remarkably, when these deep-layer methods are inserted into the updated model, their performance can even be lower than that of the zero-shot CLIP.

\noindent
\textbf{Why are shallow-layer methods superior to deep-layer ones?}
We delve deeper into the reasons why shallow-layer methods outperform deep-layer methods, investigating whether this superiority stems from the methods or the specific insertion positions.
To begin with, as illustrated in Figure~\ref{fig:preliminary} (d), we conduct experiments by training the CoOp~\citep{zhou2022learning} across various layers of CLIP.
The results reveal that, even when using the same method, the reused accuracy experiences a continuous decline with the deepening of layers, eventually falling below the zero-shot performance. 
The results align with deep-layer methods, indicating a correlation between the compatibility of the method and the layer at which it is integrated into CLIP.
Subsequently, we analyze the differences in the text encoder of ViT-B/16-based CLIP and EVA-CLIP. 
In Figure~\ref{fig:preliminary} (b-c), we calculate the average absolute and relative changes in parameters and output features across different layers. 
The results reveal that, after the model upgrade, shallow layers exhibit smaller changes in parameters and output features compared to deep layers.
This stability facilitates a smoother integration of modules into the updated model.

\subsection{Class-conditioned Context Optimization}
We propose a novel approach, \textbf{C}lass-c\textbf{on}di\textbf{t}ioned \textbf{Co}ntext \textbf{Op}timization (\textbf{ContCoOp}), designed for enhancing compatibility in VLMs.
Our approach exhibits strong compatibility in the context of model upgrades within the VLMs across various datasets.

The design of our method is rooted in two key observations. 
1) First,  shallow-layer methods exhibit superior compatibility in upgraded VLMs.
Thus, following previous works~\citep{zhou2022learning,zhou2022conditional,yao2023visual, wang2023improving}, we integrate learnable prompts at the initial layer of the text encoder in VLMs rather than designing modules at deeper layers of the model.
2) After upgrades, there is an improvement in the model's zero-shot performance.
Consequently, we posit that by leveraging zero-shot prompts as conditions, our method can dynamically adapt to model updates (due to zero-shot prompts embedding change with the model updates).
Since our approach stands on the shoulder of zero-shot prompts, it should yield superior compatibility.

The overview of our method is shown in Figure~\ref{fig:architecture}.
We provide a detailed description of our approach below.
Formally, let $\{w_i(P_i)\}_{i=1}^C$ represents the classifier weights generated by the text encoder where $w_i(P_i) = T(P_i), i=1,\dots, C$.
Here, $C$ signifies the number of classes. 
The prompt embedding is represented by $P_i = [p, c_i]\in\mathbb{R}^{(N+1)\times D}$, where $c_i$ corresponds to the embedding of the $i$-th class name, and $p = [p_1, \dots, p_N]\in\mathbb{R}^{N\times D}$ denote the learnable prompts within the embedding space. 

To enhance the compatibility of our method, we design class-conditioned prompts as $p_i(c) = p_i + Attn(p, c), i=1,\dots, N$.
Here, $c$ is the class embedding and $Attn$ denotes the attention network, which fuses the class information into prompts through the following process:
\begin{align}
\begin{aligned}
    Q = X&W_Q, \quad K = XW_K, \quad V = XW_V, \\ 
    O &= \text{Softmax}(\frac{QK^T}{\sqrt{D}})VW_O.
\end{aligned}
\end{align}
Here, $X = [p, c]\in\mathbb{R}^{(N+1)\times D}$. 
The projection weights $W_Q, W_K, W_V$, and $W_O$ correspond to query, key, value, and output, respectively, in line with standard attention networks.
After fusion, we extract the output prompts as $Attn(p, c) = O[:N]\in\mathbb{R}^{N\times D}$. 
Subsequently, these class-conditioned prompts of each class, $p(c_i) = p + Attn(p, c_i)$, are fed into the text encoder to generate the final classifier, i.e., $w_i = T([p(c_i), c_i])$.
Therefore, the classification probability can be computed as follows:
\begin{equation}
    p(y=i|x) = \frac{\exp(\cos(w_i, f)/\tau)}{\sum_{j=1}^C\exp(\cos(w_j, f)/\tau)},
\end{equation}
where $f = I(x)$ is the image feature, and $\tau=0.01$ is the temperature coefficient, and $\cos(\cdot, \cdot)$ denotes the cosine similarity.
Furthermore, if we have a zero-shot classifier, we employ the knowledge distillation loss $L_{kd}$ to fuse its knowledge into the learnable prompts as in~\citep{yao2023visual},
\begin{equation}
    L_{kd} = \frac{1}{C}\sum_{i=1}^C\|w_i - w_{zs, i}\|_2,
\end{equation}
where $w_{zs}\in\mathbb{R}^{C\times D}$ is the zero-shot classifier. 
This approach differs from deep-layer methods, as it integrates the knowledge into the input prompts in the shallow layer, enhancing its compatibility rather than simply ensembling two classifiers.
During fine-tuning, we maintain the VLM froze and optimize the learnable prompts and the attention network, guided by the loss function:
\begin{equation}
    L = L_{ce} + \lambda L_{kd},
\end{equation}
where $L_{ce}$ is the cross-entropy loss, and $\lambda$ is a hyper-parameter to control the strength of $L_{kd}$.

\begin{table}[htbp]
  \centering
  \caption{Detailed statistics of datasets used in experiments.}
  \resizebox{0.85\textwidth}{!}{
    \begin{tabular}{lrrrcc}
    \toprule
    Dataset & \# Classes & \# Training & \# Test & \multicolumn{2}{c}{Task} \\
    \midrule
    {OxfordPets} & 37    & 2,944 & 3,669 & \multicolumn{2}{c}{fine-grained pets recognition} \\
    {Flowers102} & 102   & 4,093 & 2,463 & \multicolumn{2}{c}{fine-grained flowers recognition} \\
    {FGVCAircraft} & 100   & 3,334 & 3,333 & \multicolumn{2}{c}{fine-grained aircraft recognition} \\
    {DTD} & 47    & 2,820 & 1,692 & \multicolumn{2}{c}{Textural recognition} \\
    {EuroSAT} & 10    & 13,500 & 8,100 & \multicolumn{2}{c}{Satellite image recognition} \\
    {StanfordCars} & 196   & 6,509 & 8,041 & \multicolumn{2}{c}{Fine-grained car recognition} \\
    {Food101} & 101   & 50,500 & 30,300 & \multicolumn{2}{c}{Fine-grained food recognition} \\
    {Sun397} & 397   & 15,880 & 19,850 & \multicolumn{2}{c}{Scene recognition} \\
    {Caltech101} & 100   & 4,128 & 2,465 & \multicolumn{2}{c}{Object recognition} \\
    {UCF101} & 101   & 7,639 & 3,783 & \multicolumn{2}{c}{Action recognition} \\
    {ImageNet} & 1,000 & 1.28M & 50,000 & \multicolumn{2}{c}{Object recognition} \\
    \midrule
    {ImageNetV2} & 1,000 & -     & 10,000 & \multicolumn{2}{c}{Robustness of collocation} \\
    {ImageNet-Sketch} & 1,000 & -     & 50,889 & \multicolumn{2}{c}{Robustness of sketch domain} \\
    {ImageNet-A} & 200   & -     & 7,500 & \multicolumn{2}{c}{Robustness of adversarial} \\
    {ImageNet-R} & 200   & -     & 30,000 & \multicolumn{2}{c}{Robustness of rendition styles} \\
    \bottomrule
    \end{tabular}%
    }
  \label{tab:datasets}%
\end{table}%

\section{Experiments}
\subsection{Setup}
\noindent
\textbf{Datasets.}
In accordance with previous works~\citep{radford2021learning, zhou2022learning, zhou2022conditional, wang2023improving}, we conduct the experiments across 11 publicly available datasets, encompassing a diverse set of image recognition tasks. 
These tasks include generic object recognition using ImageNet~\citep{deng2009imagenet} and Caltech101~\citep{fei2004learning}, fine-grained image recognition with OxfordPets~\citep{parkhi2012cats}, StanfordCars~\citep{krause20133d}, Flowers102~\citep{nilsback2008automated}, Food101~\citep{bossard2014food}, and FGVCAircraft~\citep{maji2013fine}. 
Additionally, we explored satellite image classification through EuroSAT~\citep{helber2019eurosat}, action classification with UCF101~\citep{soomro2012ucf101}, texture classification using DTD~\citep{cimpoi2014describing}, and scene recognition with SUN397~\citep{xiao2010sun}.
To assess the out-of-distribution generalization of our method, we further include four datasets: ImageNetV2~\citep{recht2019imagenet}, ImageNet-Sketch~\citep{wang2019learning}, ImageNet-A~\citep{hendrycks2021natural}, and ImageNet-R~\citep{hendrycks2021many}.

\noindent
\textbf{Training Details.}
To assess the module compatibility with upgraded models in vision-language models, we utilize ViT-B/16-based CLIP by default.
We train the plug-in modules on CLIP~\citep{radford2021learning} and straightforwardly integrate them into its corresponding upgraded model, EVA-CLIP~\citep{sun2023eva}.
The length of prompts is set to 16 by default.
The attention network is composed of a single multi-head attention layer with only one head, whose parameters are initialized using the Kaiming initialization method~\citep{he2015delving}.
The hyper-parameter $\lambda$ is set to 1.0 by default.
All experiments are conducted on a single NVIDIA GeForce RTX 3090.
To obtain a reliable estimate of model performance, we conduct three runs with different random seeds and average the results.

\setlength{\tabcolsep}{4pt}
\begin{table*}[t!h]
  \centering
  \captionof{table}{
    \textbf{Results in the module compatibility setting on 11 datasets.}
    We train these efficient modules using the ViT-B/16-based CLIP on 16-shot datasets.
    Subsequently, we directly integrate them into the corresponding version of EVA-CLIP.
    For comparison, we selected seven baselines: zero-shot CLIP, Linear Probe, CLIP-Adapter, Tip-Adapter, CoOp, CoCoOp, and KgCoOp.
    The evaluation metrics include the mean average of the dataset using both the original model (Base) and the upgraded model (New), along with their harmonic mean (H).
}
  \resizebox{0.9\textwidth}{!}{
    \begin{tabular}{lc|ccccccc|c}
    \toprule
    \multicolumn{2}{l|}{\textbf{Dataset}} & \textbf{ZS-CLIP} & \textbf{LP}    & \textbf{CLIP-A} & \textbf{Tip-A} & \textbf{CoOp}  & \textbf{CoCoOp} & \textbf{KgCoOp} & \textbf{ContCoOp} \\
    \midrule
    \rowcolor{gray!40}
    \multicolumn{1}{l}{\textbf{Average}} & Base  & 65.51  & 78.15  & 79.56  & \textbf{81.63}  & 79.94  & 76.27  & 78.30  & \underline{81.27} (\textcolor[RGB]{199,93,171}{\textbf{-0.36}})  \\
    \rowcolor{gray!40}
    \multicolumn{1}{l}{\hspace{3pt}\textbf{over 11}} & New   & 70.79  & 1.83  & 67.30  & 69.63  & 75.40  & 75.39  & \underline{75.77}  & \textbf{79.32} (\textcolor[RGB]{0, 155, 158}{\textbf{+3.57}})  \\
    \rowcolor{gray!40}
    \multicolumn{1}{l}{\textbf{datasets}} & H     & 68.04  & 3.57  & 72.92  & 75.15  & \underline{77.60}  & 75.83  & 77.01  & \textbf{80.28} (\textcolor[RGB]{0, 155, 158}{\textbf{+2.68}})  \\
    \midrule
    \multicolumn{1}{l}{} & Base  & 68.80  & 58.16  & 70.90  & \textbf{73.73}  & 71.62  & 71.52  & 71.04  & \underline{73.12}  \\
    \multicolumn{1}{l}{ImageNet} & New   & 74.77  & 0.09  & 62.08  & 74.52  & 75.08  & \underline{75.86}  & 75.65  & \textbf{76.50}  \\
    \multicolumn{1}{l}{} & H     & 71.66  & 0.18  & 66.18  & \underline{74.12}  & 73.31  & 73.63  & 73.27  & \textbf{74.77}  \\
    \midrule
    \multicolumn{1}{l}{} & Base  & 93.31  & 95.40  & \underline{95.92}  & 95.86  & 95.48  & 95.08  & 95.29  & \textbf{96.06}  \\
    \multicolumn{1}{l}{Caltech101} & New   & 97.16  & 0.54  & 96.15  & 96.62  & 96.74  & 97.34  & \underline{97.37}  & \textbf{97.51}  \\
    \multicolumn{1}{l}{} & H     & 95.19  & 1.08  & 96.03  & 96.24  & 96.11  & 96.19  & \underline{96.33}  & \textbf{96.78}  \\
    \midrule
    \multicolumn{1}{l}{} & Base  & 89.04  & 88.09  & 92.18  & \underline{92.87}  & 91.21  & 92.38  & \textbf{93.19}  & 92.63  \\
    \multicolumn{1}{l}{OxfordPets} & New   & 92.23  & 2.96  & 89.35  & 91.94  & 91.09  & \underline{92.92}  & \textbf{93.17}  & 92.45  \\
    \multicolumn{1}{l}{} & H     & 90.61  & 5.72  & 90.74  & 92.40  & 91.15  & \underline{92.65}  & \textbf{93.18}  & 92.54  \\
    \midrule
    \multicolumn{1}{l}{} & Base  & 65.51  & 80.53  & 79.77  & \underline{83.19}  & 82.84  & 76.60  & 76.17  & \textbf{84.16}  \\
    \multicolumn{1}{l}{StanfordCars} & New   & 79.16  & 0.29  & 75.34  & 75.31  & 80.10  & \underline{81.74}  & 81.44  & \textbf{84.01}  \\
    \multicolumn{1}{l}{} & H     & 71.69  & 0.57  & 77.49  & 79.03  & \underline{81.45}  & 79.08  & 78.72  & \textbf{84.08}  \\
    \midrule
    \multicolumn{1}{l}{} & Base  & 70.73  & \underline{97.74}  & 96.48  & 97.60  & 97.17  & 95.10  & 94.21  & \textbf{97.88}  \\
    \multicolumn{1}{l}{Flowers102} & New   & 75.80  & 0.96  & 73.99  & 74.73  & 90.23  & \underline{90.62}  & 87.18  & \textbf{96.43}  \\
    \multicolumn{1}{l}{} & H     & 73.18  & 1.90  & 83.75  & 84.65  & \underline{93.57}  & 92.81  & 90.56  & \textbf{97.15}  \\
    \midrule
    \multicolumn{1}{l}{} & Base  & 85.90  & 83.22  & 86.09  & \textbf{87.44}  & 84.18  & 86.74  & 87.23  & \underline{87.24}  \\
    \multicolumn{1}{l}{Food101} & New   & 86.56  & 1.28  & 83.84  & 86.13  & 84.05  & 86.68  & \textbf{86.89}  & \underline{86.84}  \\
    \multicolumn{1}{l}{} & H     & 86.23  & 2.51  & 84.94  & 86.78  & 84.11  & 86.71  & \textbf{87.06}  & \underline{87.04}  \\
    \midrule
    \multicolumn{1}{l}{} & Base  & 24.81  & \underline{46.42}  & 40.90  & \textbf{47.83}  & 44.42  & 37.64  & 37.34  & 42.89  \\
    \multicolumn{1}{l}{FGVCAircraft} & New   & 24.66  & 0.67  & 23.61  & 21.75  & 27.36  & \underline{30.45}  & 28.37  & \textbf{32.96}  \\
    \multicolumn{1}{l}{} & H     & 24.74  & 1.32  & 29.93  & 29.86  & \underline{33.87}  & 33.66  & 32.24  & \textbf{37.28}  \\
    \midrule
    \multicolumn{1}{l}{} & Base  & 62.56  & 69.69  & 74.78  & \textbf{76.55}  & 74.01  & 73.97  & 73.70  & \underline{76.54}  \\
    \multicolumn{1}{l}{SUN397} & New   & 70.82  & 0.30  & 62.75  & 69.92  & 73.96  & \underline{75.40}  & 74.72  & \textbf{76.26}  \\
    \multicolumn{1}{l}{} & H     & 66.44  & 0.59  & 68.23  & 73.08  & 73.99  & \underline{74.68}  & 74.21  & \textbf{76.40}  \\
    \midrule
    \multicolumn{1}{l}{} & Base  & 44.09  & 71.30  & 71.18  & 71.24  & 69.78  & 58.98  & 69.86  & \textbf{71.95}  \\
    \multicolumn{1}{l}{DTD} & New   & 50.18  & 1.77  & 47.91  & 49.41  & 65.41  & 59.40  & 67.81  & \textbf{72.08}  \\
    \multicolumn{1}{l}{} & H     & 46.94  & 3.45  & 57.26  & 58.34  & 67.52  & 59.18  & 68.81  & \textbf{72.01}  \\
    \midrule
    \multicolumn{1}{l}{} & Base  & 48.35  & 86.43  & 82.95  & \underline{86.87}  & 85.56  & 72.95  & 81.11  & \textbf{86.88}  \\
    \multicolumn{1}{l}{EuroSAT} & New   & 58.16  & 10.02  & 58.57  & 56.71  & \underline{69.05}  & 63.99  & 64.95  & \textbf{77.17}  \\
    \multicolumn{1}{l}{} & H     & 52.80  & 17.93  & 68.64  & 68.61  & \underline{76.42}  & 68.16  & 72.12  & \textbf{81.64}  \\
    \midrule
    \multicolumn{1}{l}{} & Base  & 67.46  & 82.68  & 84.04  & \textbf{84.76}  & 83.04  & 78.06  & 82.18  & \underline{84.62}  \\
    \multicolumn{1}{l}{UCF101} & New   & 69.15  & 1.23  & 66.67  & 68.86  & \underline{76.36}  & 74.93  & 75.87  & \textbf{80.32}  \\
    \multicolumn{1}{l}{} & H     & 68.30  & 2.42  & 74.35  & 75.99  & \underline{79.56}  & 76.45  & 78.90  & \textbf{82.42}  \\
    \bottomrule
    \end{tabular}%
 }
  \label{tab:model_upgrade}%
\end{table*}%

\noindent
\textbf{Evaluation Protocol.}
For module compatibility for upgraded models, we randomly select 16 samples for each dataset and train the models with 16-shot datasets using CLIP~\citep{radford2021learning}. 
Subsequently, we assess the performance on the full test dataset using both CLIP and EVA-CLIP.
The reported metric is the mean accuracy of the test dataset.
In this context, we denote the performance achieved with CLIP as the base accuracy, while the performance with EVA-CLIP is considered the new accuracy.
The harmonic mean of these two metrics is then reported as the trade-off between base and new accuracy.

\subsection{Main Results}
In this part, we present our approach and the performance of seven baselines under the module compatibility setting of VLMs. 
All methods are trained on 16-shot datasets using the ViT-B/16 CLIP.

\noindent
\textbf{Baselines.}
We evaluate our proposed approach against seven baseline methods for comparison: 
1) \textbf{Zero-Shot CLIP}~\citep{radford2021learning}: In the Zero-Shot CLIP (ZS-CLIP) baseline, we employ a prompt template such as ``a photo of a [class]" to create a zero-shot classifier, assessing its reusability.
2) \textbf{Linear Probe}~\citep{radford2021learning}: Following CLIP~\citep{radford2021learning}, we train a linear classifier on top of the pre-trained CLIP image encoder.
3) \textbf{CLIP-Adapter}~\citep{gao2023clip}: CLIP-Adapter (CLIP-A) proposes to train a task-specific adapter to adjust the visual representations.
4) \textbf{Tip-Adapter-F}~\citep{zhang2022tip}: Tip-Adapter-F (Tip-A) utilizes a cache of training data to construct the adapter, followed by fine-tuning on downstream tasks.
5) \textbf{CoOp}~\citep{zhou2022learning}: CoOp proposes learning context prompts for the input of the text encoder in downstream tasks through back-propagation.
For comparison, we choose the best version of CoOp which has 16 learnable prompts.
6) \textbf{CoCoOp}~\citep{zhou2022conditional}: CoCoOp, a variant of CoOp, employs a meta-net to fuse visual features into learnable prompts, addressing the base-to-new generalization problem.
7) \textbf{KgCoOp}~\citep{yao2023visual}: KgCoOp introduces a knowledge distillation loss to enhance CoOp's generalization ability.

\noindent
\textbf{Results.}
Table~\ref{tab:model_upgrade} illustrates the performance of our method and the baselines within the context of the model upgrade scenario.
For clarity, we highlight the highest achieved results in bold, while the second-highest results are underscored to provide a clear distinction. 
From the results, we notice that despite the deep-layer methods (LP, CLIP-A, Tip-A) have demonstrated superior performance on the original CLIP (Base), their compatibility diminishes significantly in the upgraded model, falling even below the performance of the zero-shot CLIP.
Post the model upgrade, LP, CLIP-A, and Tip-A exhibit reductions of 68.96\%, 3.49\%, and 1.16\%, respectively, when compared to the zero-shot CLIP.
This observation underscores their inability to reuse in the updated model.
In comparison with these baselines, our method outperforms, achieving the highest results on 9 out of 11 datasets when applied to the upgraded model. 
On average over 11 datasets, our method attains the highest results for New and H scores, yielding improvements of 3.57\% and 2.68\%, respectively.
This underscores the effectiveness of our approach in the context of model upgrades.

\subsection{Out-of-distribution Generalization}
We further conduct experiments to assess the robustness of our method against out-of-distribution generalization.
Specifically, we train our model utilizing the ViT-B/16-based CLIP on 16-shot ImageNet~\citep{deng2009imagenet}.
Subsequently, we transfer the model directly to target datasets, which included ImageNetV2~\citep{recht2019imagenet}, ImageNet-Sketch~\citep{wang2019learning}, ImageNet-A~\citep{hendrycks2021natural}, and ImageNet-R~\citep{hendrycks2021many}.

As presented in Table~\ref{tab:robustness}, we choose ZS-CLIP~\citep{radford2021learning}, CoOp~\citep{zhou2022learning}, CoCoOp~\citep{zhou2022conditional}, and KgCoOp~\citep{yao2023visual} for comparison.
Our method achieves the highest results on average over the four target datasets.
Specifically, our approach surpasses ZS-CLIP, CoOp, CoCoOp, and KgCoOp by 3.21\%, 1.98\%, 0.48\%, and 0.12\%, respectively, on average.
These results underscore the superior efficacy of our model in addressing challenges related to out-of-distribution generalization and mitigating the risk of overfitting on the source dataset.

\subsection{Different Architecture}
To demonstrate the adaptability of our approach, we evaluated its compatibility with an alternative CLIP architecture.
We train our method and the baseline methods on a large CLIP architecture, ViT-L/14-based CLIP.
All the methods are trained under 16-shot datasets.

The detailed results are shown in Table~\ref{tab:vitl}.
Similar to the results on ViT-B/16 architecture, our method achieves the highest results on New and H.
In comparison to the second-highest performing method, our approach achieves a 0.45\% increase on the New and a 1.03\% increase on the H.
The results demonstrate the effectiveness of our method across different CLIP architectures.


\subsection{Ablation Study}
\noindent
\begin{minipage}{0.51\textwidth}
\setlength{\tabcolsep}{4pt}
  \centering
  \captionof{table}{\textbf{Results of out-of-distribution generalization.}}
  \resizebox{1.0\textwidth}{!}{
    \begin{tabular}{llcccccc}
    \toprule
    \multicolumn{2}{c}{\multirow{2}[4]{*}{Method}} & Source & \multicolumn{4}{c}{Target}    &  \\
    \cmidrule(r){3-3} \cmidrule(r){4-8}
    \multicolumn{2}{c}{} & ImageNet & -V2   & -Sketch & -A    & -R    & Avg. \\
    \midrule
    \multicolumn{2}{l}{ZS-CLIP} & 66.73  & 60.83  & 46.15  & 47.77  & 73.96  & 57.18  \\
    \multicolumn{2}{l}{CoOp} & 71.92 & 64.18  & 46.71  & 48.41  & 74.32  & 58.41  \\
    \multicolumn{2}{l}{CoCoOp} & 71.02  & 64.07  & 48.75  & 50.63  & 76.18  & 59.91  \\
    \multicolumn{2}{l}{KgCoOp} & 71.20  & 64.10  & \textbf{48.97} & \textbf{50.69}  & 76.70  & 60.12  \\
    \rowcolor{gray!40}
    \multicolumn{2}{l}{ContCoOp} & \textbf{73.12}  & \textbf{65.50} & 48.36  & 50.43 & \textbf{77.29} & \textbf{60.39} \\
    \bottomrule
    \end{tabular}%
 }
  \label{tab:robustness}%
\end{minipage}
\hspace{0.02\textwidth}
\begin{minipage}{0.49\textwidth}
\setlength{\tabcolsep}{8pt}
  \centering
  \captionof{table}{
  Ablation of the attention network and knowledge distillation loss in our approach on the DTD dataset.
}
  \resizebox{0.9\textwidth}{!}{
    \begin{tabular}{cc|ccc}
    \toprule
    Attn  & KD    & Base  & New   & H \\
    \midrule
    \ding{55}     & \ding{55}     & 69.78  & 65.41  & 67.52  \\
    \ding{51}     & \ding{55}     & 70.27  & 69.09  & 69.67  \\
    \ding{55}     & \ding{51}     & 70.63  & 65.88  & 68.15  \\
    \ding{51}     & \ding{51}     & \textbf{71.95}  & \textbf{72.08}  & \textbf{72.01}  \\
    \bottomrule
    \end{tabular}%
 }
  \label{tab:component}%
\end{minipage}

\vspace{10pt}
\noindent
\textbf{Effectiveness of different components.}
Our method is composed of two pivot components: the attention network (Attn) and the knowledge distillation (KD) loss, compared with previous work CoOp.
In this part, we ablate the effectiveness of these components.
The results are presented in Table~\ref{tab:component}, 
The results demonstrate that integrating the attention network and KD loss independently yields enhanced module reusability.
Notably, it is observed that the advantages stemming from the attention network surpass those of KD, showcasing a more pronounced improvement. Furthermore, combining these two components produces the best performance, yielding the effectiveness of our method.

\noindent
\textbf{Impact of variable context lengths.}
We analyze the influence of varying context lengths for prompts in our proposed method, as illustrated in Figure~\ref{fig:context_length}. 
Our experiments involve training our model with context lengths of 4, 8, and 16 on the UCF101 dataset. 
Notably, our findings reveal that the prompt length has a negligible impact on the performance of our method.
Given that a context length of 16 yielded the optimal results, we adopt a context length of 16 for training our method, ensuring the attainment of superior results.

\noindent
\begin{minipage}{0.48\textwidth}
\setlength{\tabcolsep}{6pt}
  \centering
  \captionof{table}{Results of ContCoOp with different $\lambda$.}
  \resizebox{1.0\textwidth}{!}{
    \begin{tabular}{c|ccccc}
    \toprule
       $\lambda$   & 0.00  & 0.30  & 0.50  & 0.70  & 1.00  \\
    \midrule
    Base  & 83.66  & 84.80  & 84.62  & 84.88  & \textbf{85.19} \\
    New   & 79.26  & 79.77  & \textbf{80.32} & 79.90  & 80.18  \\
    H     & 81.40  & 72.21  & 82.42  & 82.31  & \textbf{82.61} \\
    \bottomrule
    \end{tabular}%
 }
  \label{tab:lambda}%
\end{minipage}
\hspace{0.02\textwidth}
\begin{minipage}{0.48\textwidth}
  \centering
  \setlength{\tabcolsep}{6pt}
  \captionof{table}{Results of our method with different heads.}
  \resizebox{1.0\textwidth}{!}{
    \begin{tabular}{c|ccccc}
    \toprule
       \#heads   & 1  & 2  & 4  & 8  & 16  \\
    \midrule
    Base  & 71.95  & \textbf{72.16}  & 72.04  & 72.08  & 72.06  \\
    New   & \textbf{72.08}  & 71.95  & 71.73  & 71.04  & 71.22  \\
    H     & \textbf{72.01}  & 72.00  & 71.88  & 71.56  & 71.64  \\
    \bottomrule
    \end{tabular}%
 }
  \label{tab:number_heads}%
\end{minipage}

\noindent
\begin{minipage}{0.48\textwidth}
\setlength{\tabcolsep}{6pt}
  \centering
  \captionof{table}{Different condition information of our method.}
  \resizebox{1.0\textwidth}{!}{
    \begin{tabular}{cc|ccc}
    \toprule
    \multicolumn{2}{c|}{} & Base  & New   & H \\
    \midrule
    \multicolumn{2}{c|}{[class]} & 71.95  & \textbf{72.08}  & \textbf{72.01}  \\
    \multicolumn{2}{c|}{a photo of a [class]} & \textbf{72.01}  & 70.74  & 71.37  \\
    \multicolumn{2}{c|}{[class] texture} & 71.51  & 71.53  & 71.52  \\
    \bottomrule
    \end{tabular}%
 }
  \label{tab:condition}%
\end{minipage}
\hspace{0.02\textwidth}
\begin{minipage}{0.40\textwidth}
    \centering
    \includegraphics[width=1.0\textwidth]{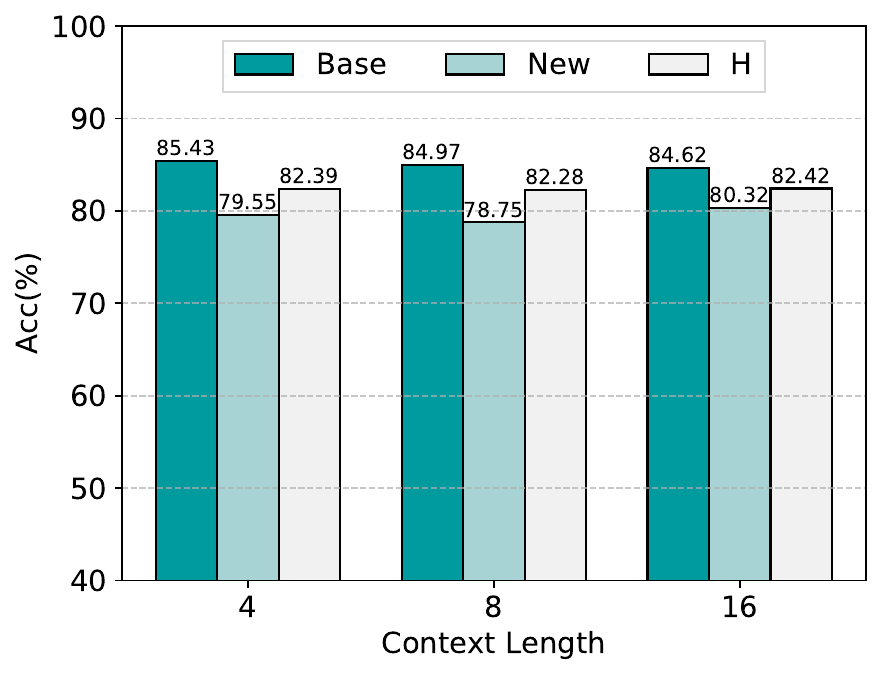}
    \captionof{figure}{The impact of different context length.}
    \label{fig:context_length}
\end{minipage}

\noindent
\textbf{Sensitivity analysis of the hyper-parameter $\lambda$.}
In our method, there is a hyper-parameter $\lambda$ that controls the strength of knowledge distillation loss.
To assess its impact, we conduct a sensitivity analysis on the UCF101 dataset.
The results are shown in Table~\ref{tab:lambda}.
We observe that the optimal performance is attained when $\lambda$ is set to 1. 
Consequently, we adopt $\lambda=1$ by default for all datasets.

\noindent
\textbf{Impact of varying the number of heads in the attention network.}
Our method incorporates an attention network with a self-attention layer.
In this part, we explore the influence of adjusting the number of heads within the attention network.
To assess the impact, we conducted experiments by training our model with varying numbers of heads on the DTD dataset.
The results are shown in Table~\ref{tab:number_heads}.
The results indicate that employing a smaller number of attention heads yields better performance.
Consequently, we opt to maintain a single head in the attention network.

\noindent
\textbf{Impact of conditional information.}
In ContCoOp, we leverage learnable prompts conditioned on class embeddings. 
However, it is noteworthy that our method allows for longer prompts as conditions, such as ``a photo of a [class]".
To ablate its influence, we train our method with different conditioned information on the DTD dataset.
The results are shown in Table~\ref{tab:condition}.
Remarkably, our analysis reveals that opting exclusively for class embeddings as conditions yields the best results.
This observation can be attributed to the embedding level, where prompts such as ``a photo of a [class]" may lack significant semantic information. 
Due to the training methodology of CLIP, the semantic information of the prompt is primarily encoded in text features. 
Consequently, employing longer prompts introduces unnecessary noise during training, potentially hampering the performance.

\setlength{\tabcolsep}{6pt}
\begin{table}[t!h]
  \centering
  \caption{
  \textbf{Results in the module compatibility setting on ViT-L/14 architecture.}
    We train these efficient modules using the ViT-L/14-based CLIP on 16-shot datasets.
    Subsequently, we directly integrate them into the corresponding version of EVA-CLIP.
    For comparison, we selected seven baselines: zero-shot CLIP, Linear Probe, CLIP-Adapter, Tip-Adapter, CoOp, CoCoOp, and KgCoOp.
    The evaluation metrics include the mean average of the dataset using both the original model (Base) and the upgraded model (New), along with their harmonic mean (H).
 }
  \resizebox{0.9\textwidth}{!}{
    \begin{tabular}{llc|ccccccc|c}
    \toprule
    \multicolumn{3}{l|}{Dataset} & ZS-CLIP & LP    & CLIP-A & Tip-A & CoOp  & CoCoOp & KgCoOp & ContCoOp \\
    \midrule
    \multicolumn{2}{l}{\multirow{3}[1]{*}{Average}} & Base  & 72.76  & 82.74  & 84.86  & \textbf{85.90} & 84.66  & 83.71  & 83.61  & \underline{85.34} (\textcolor[RGB]{199,93,171}{\textbf{-0.56}})  \\
    \multicolumn{2}{l}{} & New   & 77.07  & 2.34  & 74.54  & 75.38  & 80.96  & \underline{82.47}  & 80.23  & \textbf{82.92} (\textcolor[RGB]{0, 155, 158}{\textbf{+0.45}}) \\
    \multicolumn{2}{l}{} & H     & 74.85  & 4.56  & 79.37  & 80.30  & \underline{82.77}  & 83.08  & 81.89  & \textbf{84.11} (\textcolor[RGB]{0, 155, 158}{\textbf{+1.03}}) \\
    \midrule
    \multicolumn{2}{l}{\multirow{3}[1]{*}{ImageNet}} & Base  & 75.95  & 65.74  & 77.04  & \textbf{79.54} & 78.02  & 77.73  & 77.56  & \underline{78.99}  \\
    \multicolumn{2}{l}{} & New   & 79.88  & 0.10  & 71.33  & 79.62  & 80.07  & \underline{80.54}  & 80.44  & \textbf{81.04} \\
    \multicolumn{2}{l}{} & H     & 77.87  & 0.20  & 74.08  & \underline{79.58}  & 79.03  & 79.11  & 78.97  & \textbf{80.00} \\
    \midrule
    \multicolumn{2}{l}{\multirow{3}[1]{*}{Caltech101}} & Base  & 95.17  & 96.92  & 97.40  & \textbf{97.69} & 97.23  & 97.19  & \underline{97.61}  & 97.55  \\
    \multicolumn{2}{l}{} & New   & 97.57  & 0.70  & 97.23  & 97.16  & 97.59  & \textbf{98.15} & 98.03  & \underline{98.07}  \\
    \multicolumn{2}{l}{} & H     & 96.35  & 1.40  & 97.32  & 97.42  & 97.41  & 97.67  & \textbf{97.82} & \underline{97.81}  \\
    \midrule
    \multicolumn{2}{l}{\multirow{3}[1]{*}{OxfordPets}} & Base  & 93.43  & 92.01  & 94.48  & 94.85  & 93.99  & \underline{95.04}  & \textbf{95.10} & 94.62  \\
    \multicolumn{2}{l}{} & New   & 93.92  & 3.19  & 92.47  & 93.74  & 93.50  & \textbf{94.59} & 94.03  & \underline{94.11}  \\
    \multicolumn{2}{l}{} & H     & 93.68  & 6.15  & 93.46  & 94.29  & 93.74  & \textbf{94.81} & \underline{94.56}  & 94.37  \\
    \midrule
    \multicolumn{2}{l}{\multirow{3}[1]{*}{StanfordCars}} & Base  & 76.84  & 85.74  & 87.55  & \underline{88.89}  & 88.77  & 86.02  & 84.54  & \textbf{89.43} \\
    \multicolumn{2}{l}{} & New   & 90.09  & 0.50  & 87.81  & 89.52  & 89.00  & \textbf{91.20} & \underline{90.88}  & 90.64  \\
    \multicolumn{2}{l}{} & H     & 82.94  & 1.00  & 87.68  & \underline{89.20}  & 88.88  & 88.53  & 87.60  & \textbf{90.03} \\
    \midrule
    \multicolumn{2}{l}{\multirow{3}[1]{*}{Flowers102}} & Base  & 79.46  & \textbf{99.34} & 98.75  & 99.00  & \underline{99.04}  & 97.90  & 96.67  & 98.90  \\
    \multicolumn{2}{l}{} & New   & 77.22  & 0.84  & 77.44  & 70.42  & \underline{94.51}  & 93.00  & 85.93  & \textbf{94.94} \\
    \multicolumn{2}{l}{} & H     & 78.32  & 1.66  & 86.81  & 82.29  & \underline{96.72}  & 95.38  & 90.98  & \textbf{96.88} \\
    \midrule
    \multicolumn{2}{l}{\multirow{3}[1]{*}{Food101}} & Base  & 90.91  & 89.47  & 91.13  & \textbf{91.82} & 90.10  & 91.22  & \underline{91.64}  & 91.61  \\
    \multicolumn{2}{l}{} & New   & 91.02  & 1.07  & 90.11  & 91.01  & 89.59  & 90.88  & \underline{91.33}  & \textbf{91.38} \\
    \multicolumn{2}{l}{} & H     & 90.97  & 2.12  & 90.62  & 91.41  & 89.84  & 91.05  & \underline{91.48}  & \textbf{91.49} \\
    \midrule
    \multicolumn{2}{l}{\multirow{3}[1]{*}{FGVCAircraft}} & Base  & 32.61  & 55.92  & 55.94  & \textbf{59.48} & \underline{56.32}  & 51.65  & 49.78  & 53.88  \\
    \multicolumn{2}{l}{} & New   & 35.94  & 1.17  & 34.00  & 33.58  & \underline{44.04}  & \textbf{46.11} & 40.65  & 43.72  \\
    \multicolumn{2}{l}{} & H     & 34.20  & 2.29  & 42.29  & 42.91  & \textbf{49.38} & 48.72  & 44.75  & \underline{48.26}  \\
    \midrule
    \multicolumn{2}{l}{\multirow{3}[1]{*}{SUN397}} & Base  & 67.66  & 72.36  & 78.34  & \textbf{79.72} & 77.26  & 77.40  & 77.04  & \underline{79.51}  \\
    \multicolumn{2}{l}{} & New   & 74.56  & 0.37  & 69.35  & 73.37  & 76.57  & \underline{78.53}  & 78.22  & \textbf{79.54} \\
    \multicolumn{2}{l}{} & H     & 70.94  & 0.74  & 73.57  & 76.41  & 76.91  & \underline{77.96}  & 77.63  & \textbf{79.53} \\
    \midrule
    \multicolumn{2}{l}{\multirow{3}[1]{*}{DTD}} & Base  & 53.01  & 75.22  & \underline{75.67}  & 75.43  & 73.44  & 71.73  & 75.24  & \textbf{76.38} \\
    \multicolumn{2}{l}{} & New   & 63.36  & 2.52  & 57.96  & 61.54  & 69.70  & \underline{72.04}  & 70.55  & \textbf{74.55} \\
    \multicolumn{2}{l}{} & H     & 57.73  & 4.87  & 65.64  & 67.78  & 71.51  & 71.88  & \underline{72.81}  & \textbf{75.45} \\
    \midrule
    \multicolumn{2}{l}{\multirow{3}[1]{*}{EuroSAT}} & Base  & 60.33  & \textbf{90.69} & 88.83  & \underline{90.48}  & 89.85  & 88.44  & 87.97  & 89.43  \\
    \multicolumn{2}{l}{} & New   & 67.43  & 14.21  & 66.64  & 62.74  & 73.42  & \underline{77.66}  & 71.95  & \textbf{78.94} \\
    \multicolumn{2}{l}{} & H     & 63.69  & 24.44  & 76.15  & 74.05  & 80.68  & \underline{82.67}  & 79.15  & \textbf{83.59} \\
    \midrule
    \multicolumn{2}{l}{\multirow{3}[1]{*}{UCF101}} & Base  & 74.99  & 86.72  & \underline{88.38}  & 88.03  & 87.30  & 86.44  & 86.57  & \textbf{88.41} \\
    \multicolumn{2}{l}{} & New   & 76.79  & 1.11  & 75.54  & 76.46  & 82.62  & \underline{84.46}  & 80.56  & \textbf{85.20} \\
    \multicolumn{2}{l}{} & H     & 75.88  & 2.19  & 81.46  & 81.83  & 84.89  & \underline{85.44}  & 83.46  & \textbf{86.77} \\
    \bottomrule
    \end{tabular}%
  }
  \label{tab:vitl}%
\end{table}%

\section{Conclusion}
In this paper, we proposed a crucial issue in efficient fine-tuning for VLMs, i.e., \textit{Are the efficient fine-tuned modules still effective to the upgraded model?}
To investigate this, we first conducted experiments on the compatible performance of existing methods. 
The results revealed that, to varying degrees, these methods exhibit limitations in terms of compatibility.
To address this issue, we proposed a novel approach, \textbf{C}lass-c\textbf{on}di\textbf{t}ioned \textbf{Co}ntext \textbf{Op}timization (\textbf{ContCoOp}).
ContCoOp generated the class-conditioned prompts through an attention network, thus the prompts can be dynamically updated for the upgraded model.
We have conducted experiments over 15 datasets, illustrating the superior performance of ContCoOp.
Moreover, ContCoOp not only demonstrated strong compatibility for model upgrades but also exhibited robust performance in out-of-distribution generalization.
In the future, we plan to enhance our method and extend our research to explore the issue of compatibility in additional modalities, such as NLP.

\section*{Acknowledgments}
This work was funded by the National Natural Science Foundation of China under Grant 62276256 and the Young Elite Scientists Sponsorship Program by CAST (2023QNRC001).

\bibliography{refs}

\end{document}